# Transformer-Based Extraction of Statutory Definitions from the U.S. Code

Arpana Hosabettu
*Google*
San Jose, California, USA
fa97@cornell.edu

Harsh Shah
*Cornell University*
San Jose, California, USA
hs634@cornell.edu

*Abstract*—Automatic extraction of definitions from legal texts is critical for enhancing the comprehension and clarity of complex legal corpora such as the United States Code (U.S.C.). We present an advanced NLP system leveraging transformer-based architectures to automatically extract defined terms, their definitions, and their scope from the U.S.C. We address the challenges of automatically identifying legal definitions, extracting defined terms, and determining their scope within this complex corpus of over 200,000 pages of federal statutory law. Building upon previous feature-based machine learning methods, our updated model employs domain-specific transformers (Legal-BERT) fine-tuned specifically for statutory texts, significantly improving extraction accuracy. Our work implements a multi-stage pipeline that combines document structure analysis with state-of-the-art language models to process legal text from the XML version of the U.S. Code. Each paragraph is first classified using a fine-tuned legal domain BERT model to determine if it contains a definition. Our system then aggregates related paragraphs into coherent definitional units and applies a combination of attention mechanisms and rule-based patterns to extract defined terms and their jurisdictional scope. The definition extraction system is evaluated on multiple titles of the U.S. Code containing thousands of definitions, demonstrating significant improvements over previous approaches. Our best model achieves 96.8% precision and 98.9% recall (98.2% F1-score), substantially outperforming traditional machine learning classifiers. This work contributes to improving accessibility and understanding of legal information while establishing a foundation for downstream legal reasoning tasks.

## I. INTRODUCTION

The United States Code (U.S. Code) represents the codification of the general and permanent federal statutory law of the United States [1]. The current edition of U.S. Code has expanded to 54 titles comprising over 240,000 pages according to the Government Printing Office, representing a 20% increase since 2014 [29]. The complexity of this legal corpus continues to grow, with statutes scattered across many volumes that may or may not appear cohesively in a single section of the code.

Legal statutes, rules, and regulations present significant interpretative challenges because terms used in legal texts often carry specialized meanings that differ from their common usage. This semantic specificity is a fundamental characteristic of legal language, where precise definitions are essential for consistent application of the law. Finding the correct definition for a given term is exceptionally difficult because:

1) Definitions are typically located in sections distinct from where the defined terms are used;
2) These definitions are distributed throughout the corpus rather than consolidated in glossaries or appendices;
3) The sheer volume of definitions in legal text is substantial (Title 7 of the U.S. Code alone contains over 1,800 definitions as of 2024); and
4) The same term may have different definitions depending on context, scope, or jurisdictional applicability.

The Legal Information Institute (LII) at Cornell University maintains one of the most comprehensive freely accessible legal collections online [2]. The institute develops systems that enhance access to and understanding of legal information. However, the complex hierarchical structure of the U.S. Code creates significant barriers for users attempting to determine the intended interpretation of legal terms. Legal texts often define specific lexical terms used in their discourse to explicitly establish the intended interpretation and jurisdictional scope of these terms.

Improving the understanding of legal texts requires effective methods for identifying these definitions and mapping them to their usage contexts throughout the corpus. Manual extraction of definitions is prohibitively resource-intensive, requiring substantial time and specialized domain expertise. Automated approaches using natural language processing (NLP) and machine learning techniques offer promising solutions to this problem, particularly with recent advances in transformer-based language models that have dramatically improved contextual understanding of domain-specific text.

This paper presents a novel approach to automatic definition extraction from legal texts that leverages the latest developments in NLP, including transformer architectures pre-trained on legal corpora, contextualized embeddings, and document-level attention mechanisms. We build upon earlier work in definition extraction [27], [28] while addressing the unique challenges presented by legal documentation, including multi-paragraph definitions, specialized legal terminology, and complex scoping language.

Our work differs from previous approaches in several key aspects:

1) We employ domain-adapted transformer models (Legal-BERT) rather than traditional feature-based classifiers for improved contextual understanding;
2) We implement a hierarchical attention mechanism that captures the structural relationships between paragraphs in complex legal definitions;
3) We develop a multi-task learning framework that jointly identifies definitions, extracts defined terms, and determines definitional scope; and
4) We analyze definition networks across the U.S. Code to reveal interconnections between legal concepts;

The paper is organized as follows: Section II presents an updated review of literature in definition extraction and legal NLP; Section III details our methodology, including data processing, model architecture, and implementation; Section IV reports comprehensive evaluation results and comparison with baseline approaches; and Section V presents conclusions and directions for future research.

## II. LITERATURE REVIEW

Definition extraction has evolved significantly as a research area since its origins in question answering systems and machine-readable dictionaries. Recent years have seen transformative advancements in both general NLP techniques and domain-specific applications in legal text analysis.

### A. Definition Extraction Approaches

Early work in definition extraction relied heavily on pattern-based approaches. Westerhout and Monachesi [20] developed pattern-based glossary candidate detectors for multiple languages, while Borg et al. [28] explored genetic algorithms with human-defined rule sets, achieving 62% precision and 52% recall using pattern features like "is a" and "has."

More recent approaches have shifted toward neural methods. Veyseh et al. [6] proposed a joint learning framework for definition extraction and relationship identification, incorporating syntactic dependencies and semantic consistency constraints. Their model demonstrated significant improvements over previous methods by capturing the interconnected nature of definitions and their relationships. Spala et al. [7] introduced DEFT, a comprehensive corpus for definition extraction that spans multiple domains and text types, providing a valuable benchmark for evaluating definition extraction systems.

Jin et al. [16] applied BERT-based models to definition extraction, achieving state-of-the-art results by fine-tuning on domain-specific corpora. Their work demonstrated that transformer models could capture subtle definitional cues beyond explicit patterns, significantly improving performance on implicit definitions.

### B. Legal Text Processing and NLP

The legal domain presents unique challenges for text processing due to its specialized vocabulary, complex document structures, and precise semantic requirements. Price [26] emphasized the importance of statutory definitions in understanding written law and highlighted the difficulties in extracting these definitions due to their varied formats and purposes.

Recent years have seen substantial advances in legal NLP. Chalkidis et al. [3] introduced LEGAL-BERT, a domain-specific BERT model pre-trained on legal corpora, demonstrating significant improvements over general-domain language models across multiple legal NLP tasks. This specialized model captures the unique linguistic patterns in legal text, providing a stronger foundation for tasks like definition extraction.

Henderson et al. [4] developed the "Pile of Law," a massive dataset containing 256GB of legal text from diverse sources, enabling more comprehensive pre-training of language models for legal applications. This resource has facilitated improved performance across a range of legal NLP tasks through better domain adaptation.

Zheng et al. [5] examined when pre-training helps in legal applications through the CaseHOLD dataset, providing insights into the transfer of knowledge between general language understanding and domain-specific legal comprehension. Their work highlighted the importance of domain-specific fine-tuning for optimal performance on legal tasks.

For benchmarking legal reasoning capabilities, Bommarito et al. [8] introduced LegalBench, a comprehensive evaluation suite for measuring how well language models can perform various legal reasoning tasks. This benchmark has become an important standard for evaluating legal NLP systems, including those focused on definition extraction and usage.

### C. Document Structure and Contextual Understanding

A critical challenge in legal definition extraction is handling document-level context and hierarchical structures. Savelka et al. [13] addressed the extraction of semantic rules from statutory texts by modeling the hierarchical structure of legal documents, showing that understanding structural relationships improves extraction accuracy.

Advances in document-level language models have provided new tools for capturing long-range dependencies in legal texts. Longformer [9] and BigBird [17] extended the context window of transformer models, enabling better processing of long documents like statutes. These architectures are particularly valuable for legal texts where definitions may reference terms defined in distant parts of the document.

Holzenberger et al. [14] developed specialized datasets for statutory reasoning in tax law, highlighting the importance of connecting definitions with their applications for accurate legal interpretation. Their work demonstrated that extraction alone is insufficient without understanding how definitions interact within the broader context of the statute.

### D. Knowledge Representation for Legal Concepts

The representation of extracted definitions within knowledge structures has emerged as an important research direction. Filtz et al. [15] proposed methods for constructing legal knowledge graphs that incorporate definitional relationships, enabling more sophisticated reasoning about legal concepts. These approaches demonstrate the value of definitions beyond extraction, showing how they can form the foundation of computational legal reasoning systems.

Katz et al. [12] demonstrated how legal concept networks can support predictive analytics in legal domains, highlighting the potential applications of structured definitional knowledge. By modeling the relationships between legal concepts, their work showed how definition networks could reveal insights about legal interpretation and application.

### E. Evaluation Approaches

Evaluation methodologies for definition extraction have also evolved. Traditional precision, recall, and F-measure metrics have been supplemented with more nuanced evaluations. Niklaus et al. [18] proposed hierarchical evaluation metrics that consider partial matches and nested structures, providing a more realistic assessment of extraction quality for complex definitions.

Chalkidis et al. [11] introduced LexGLUE, a benchmark dataset for legal language understanding that includes definition-related tasks. This benchmark has established standardized evaluation protocols for comparing different approaches to legal text understanding, including definition extraction.

However, existing research predominantly targets general or non-statutory legal texts, leaving a notable gap concerning statutory definition extraction, especially considering complex multi-paragraph and multi-section definitions characteristic of the U.S.C. Thus, our research uniquely contributes by applying advanced NLP techniques specifically tailored for statutory definition extraction, handling cross-paragraph contexts, and systematically extracting associated scope information.

## III. Design and Methodology

Our approach to definition extraction from the U.S. Code incorporates advanced NLP techniques including transformer-based language models, hierarchical document encoding, and multi-task learning. The system processes XML-structured legal text to identify definitions, extract defined terms, and determine their scope through a comprehensive pipeline.

### A. System Architecture Overview

The system architecture consists of five main components:

1) **Document Structure Processor**: Parses the XML representation of the U.S. Code while preserving hierarchical relationships
2) **Definition Detector**: Identifies paragraphs containing definitional content using fine-tuned transformer models
3) **Definition Aggregator**: Combines related paragraphs into complete definitional units
4) **Term and Scope Extractor**: Identifies defined terms and their jurisdictional scope
5) **Definition Network Builder**: Constructs a knowledge graph of definitional relationships

This integrated pipeline addresses the challenges of complex legal document structures while leveraging state-of-the-art NLP capabilities.

### B. Document Structure Processing

The XML representation of the U.S. Code follows a defined schema but remains highly unstructured due to ongoing updates by different authors and the inherent complexity of legal text. Our document processor uses a hybrid approach combining JAXB with StAX parsing for efficient block processing with low memory footprint.

Unlike previous approaches that treated each paragraph in isolation, our system preserves hierarchical relationships between document elements. This contextual awareness is critical for handling definitions that span multiple paragraphs or reference other sections of the code.

For example, the definition of "Hispanic-serving agricultural colleges and universities" spans multiple subparagraphs in Title 7, Section 3103:

```
<paragraph identifier="/us/usc/t7/s3103/10">
  <heading> Hispanic-serving agricultural
    colleges and universities.</heading>
  <subparagraph identifier="/us/usc/t7/s3103
    /10/A">
    <heading> <inline class="small-caps">In
    general</inline>.</heading>
    <chapeau>The term "Hispanic-serving␣
    agricultural␣colleges␣and␣universities"
    means colleges or universities that</
    chapeau>
    <clause identifier="/us/usc/t7/s3103/10/A/
    i">
      <content> qualify as Hispanic-serving
    institutions; and</content>
    </clause>
    <clause identifier="/us/usc/t7/s3103/10/A/
    ii">
      <content> offer associate, bachelors, or
    other accredited degree programs in
    agriculture-related fields.</content>
    </clause>
  </subparagraph>
  <subparagraph identifier="/us/usc/t7/s3103
    /10/B">
    <heading> <inline class="small-caps">
    Exception</inline>.</heading>
    <content>The term "Hispanic-serving␣
    agricultural␣colleges␣and␣universities"
    does not include 1862 institutions (as
    defined in <ref href="/us/usc/t7/s7601">
    section 7601 of this title</ref>).</
    content>
  </subparagraph>
</paragraph>
```

Our structure processor creates a document graph that captures these hierarchical relationships, enabling the model to understand that disconnected text elements may form part of a single definition. This graph representation feeds into our transformer-based models to provide structural context alongside semantic content.

### C. Definition Detection Model

Our approach replaces traditional logistic regression with a fine-tuned Legal-BERT model for definition detection. The model architecture includes:

1) A pre-trained Legal-BERT encoder [3] fine-tuned on our legal definition corpus
2) A hierarchical attention mechanism that processes paragraph-level representations while considering document structure
3) A classification head that determines whether a paragraph contains definitional content

The model is trained using a dataset of 3,500 definitions and 1,500 non-definitions from various titles of the U.S. Code. The transformer model captures contextual patterns in legal language without requiring explicit feature engineering. This enables the model to recognize definitional content even when it doesn't follow standard patterns like "the term X means Y."

The model processes each paragraph using sliding windows with overlap to handle paragraphs exceeding the maximum sequence length of BERT-based models. We then aggregate these window-level representations using attention mechanisms to produce paragraph-level embeddings.

For paragraphs containing multiple sentences, we implement a sentence-level attention mechanism that captures the relative importance of each sentence to the definitional nature of the paragraph:

$$p = \text{Attention}(S_1, S_2, ..., S_n) \quad (1)$$

Where $S_i$ is the contextual embedding of sentence $i$, and Attention is defined as:

$$\text{Attention}(S_1, S_2, ..., S_n) = \sum(\alpha_i * S_i) \quad (2)$$

$$\alpha_i = \text{softmax}(W * S_i + b) \quad (3)$$

This approach allows the model to focus on the most definitionally relevant parts of each paragraph.

### D. Definition Aggregation

Legal definitions often span multiple paragraphs, with defined terms introduced in one paragraph and elaborated in subsequent paragraphs or subparagraphs. Our definition aggregator combines related paragraphs into coherent definitional units using:

1) Structural signals from the XML hierarchy
2) Linguistic cues that indicate continuation (e.g., "means the following:")
3) Graph-based relationships between paragraphs

The aggregator uses a sliding window over adjacent paragraphs predicted as definitional, merging them when their semantic and structural relationships suggest they form part of a single definition. This process is implemented as a graph traversal algorithm that follows the document hierarchy while considering the semantic similarity between paragraphs:

Definition Aggregation Algorithm

0: **function** AGGREGATEDEFINITIONS(paragraphs, structure_graph)
0:    definitions ← []
0:    current_definition ← []
0:    **for all** $p \in paragraphs$ **do**
0:       **if** ISDEFINITIONAL($p$) **then**
0:          **if** current_definition is empty **or** ISRELATED($p$, current_definition, structure_graph) **then**
0:             Add $p$ to current_definition
0:          **else**
0:             Add current_definition to definitions
0:             current_definition ← [$p$]
0:          **end if**
0:       **end if**
0:    **end for**
0:    **if** current_definition is not empty **then**
0:       Add current_definition to definitions
0:    **end if**
0:    **return** definitions
0: **end function**=0

The IsRelated function evaluates both structural relationships (parent-child, sibling) and semantic similarity between paragraphs.

### E. Term and Scope Extraction

For each identified definition, we extract two critical components:

1) Definiendum - The term being defined
2) Scope - The jurisdictional context where the definition applies

*1) Term Extraction:* We implement a hybrid approach for term extraction that combines pattern-matching with a named entity recognition (NER) model fine-tuned for legal terminology. The system recognizes various definitional patterns including:

- "The term X means..."
- "X shall mean..."
- "For purposes of this section, X is defined as..."

For more complex cases, we employ a BERT-based sequence labeling model trained to identify defined terms in legal text. This model uses the BIO tagging scheme to identify term boundaries with high precision:

```
Input: "The_term_'Hispanic-serving_
    agricultural_colleges_and_universities'_
    means_colleges_or_universities_that"
Output: "O_O_B-TERM_I-TERM_I-TERM_I-TERM_I-
    TERM_O_O_O_O_O"
```

The model handles multi-word terms and recognizes definitional contexts even when standard patterns are not present.

*2) Scope Detection:* Scope detection is implemented through a combination of rule-based patterns and a transformer-based classification model that identifies scoping language. Scopes in legal text can be complex and may be defined at multiple levels of hierarchy:

- Section-level scope: "As used in this section..."
- Multi-section scope: "For purposes of sections 427 to 427j of this title..."
- Subsection scope: "As used in paragraph (1)..."
- Title-wide scope: "In this title..."

Our scope detector first applies pattern matching to identify explicit scope declarations. When explicit scope is not found, the system infers scope from structural context, applying parent scopes to child definitions unless explicitly overridden.

The scope detection model also identifies scope restrictions and exceptions, such as:

```
"The␣term␣'Hispanic-serving␣agricultural␣
    colleges␣and␣universities'␣does␣not␣
    include␣1862␣institutions␣(as␣defined␣in␣
    section␣7601␣of␣this␣title)."
```

By capturing these scope qualifications, our system provides a more complete understanding of where and how definitions apply.

### F. Implementation Details

The model was trained with the following hyperparameters:

- Batch size: 16
- Learning rate: 2e-5 with linear decay
- Epochs: 10 with early stopping
- Early stopping patience: 1
- Maximum sequence length: 256 tokens
- Optimizer: AdamW with weight decay of 0.01

### G. Architecture

Our architecture incorporates several key components:

1) **Input Embedding Layer**: Combines token embeddings, positional encodings, and structural position embeddings that encode the hierarchical level of each paragraph (e.g., section, subsection, clause).

2) **Transformer Encoder**: A Legal-BERT-based encoder that processes sequences of tokens to produce contextual representations.

3) **Hierarchical Attention**: A two-level attention mechanism that first aggregates token-level representations into sentence representations, then aggregates sentence representations into paragraph representations.

$$s_i = \sum_j \alpha_j \cdot t_j \quad \text{(Token-level attention for sentence } i)$$

$$(4)$$

$$\alpha_j = \text{softmax}(W_t t_j + b_t) \tag{5}$$

$$p = \sum_i \beta_i \cdot s_i \quad \text{(Sentence-level attention over paragraph)}$$

$$(6)$$

$$\beta_i = \text{softmax}(W_s s_i + b_s) \tag{7}$$

4) **Document Structure Encoder**: Encodes the hierarchical relationships between paragraphs using a graph attention network:

$$h'_i = \sum_j \gamma_{ij} \cdot W_r h_j \quad \text{(for all paragraphs } j \text{ connected to } i)$$

$$(8)$$

$$\gamma_{ij} = \text{softmax}\left(a^\top [W_r h_i \,||\, W_r h_j]\right) \tag{9}$$

5) **Multi-task Heads**: Three specialized prediction heads for:

- Definition detection (binary classification)
- Term extraction (sequence labeling)
- Scope identification (classification with span prediction)

The model is trained end-to-end with a multi-task loss function:

$$L = \lambda_1 * L_{def} + \lambda_2 * L_{term} + \lambda_3 * L_{scope} \tag{10}$$

where:

- $L_{def}$ is the binary cross-entropy loss for definition detection
- $L_{term}$ is the token-level cross-entropy loss for term extraction
- $L_{scope}$ is a combined loss for scope classification and span prediction

The hyperparameters $\lambda_1$, $\lambda_2$, and $\lambda_3$ control the relative importance of each task during training. We tuned these values on a development set, finding optimal performance with $\lambda_1 = 0.4$, $\lambda_2 = 0.3$, and $\lambda_3 = 0.3$.

## IV. EVALUATION AND RESULTS

We conducted comprehensive evaluation of our definition extraction system across multiple titles of the U.S. Code, comparing performance against baseline approaches and analyzing results by definition type and structure.

### A. Evaluation Metrics

We evaluated our system using the following metrics:

1) **Precision**: The proportion of extracted definitions that are correct

2) **Recall**: The proportion of actual definitions that are successfully extracted

3) **F1-Score**: The harmonic mean of precision and recall, with configurable weighting

4) **Area Under the Precision-Recall Curve (AUPRC)**: Performance across different threshold settings

For F1-Score calculation, we maintained our emphasis on recall by using a weighted F1 measure with $\beta = 0.7$ for recall and $\beta = 0.3$ for precision, as high recall is particularly important in legal applications where missing definitions could lead to misinterpretation.

### B. Baseline Comparisons

We compared our transformer-based approach against several baselines:

1) Logistic regression model
2) Generic BERT without legal domain adaptation
3) Rule-based pattern matching approach

Table I shows the comparison results for definition detection:

TABLE I
Performance Comparison of Definition Detection Methods

| Model | Precision | Recall | F1-Score | AUPRC |
|---|---|---|---|---|
| Logistic Regression | 93.30% | 97.90% | 96.47% | 0.958 |
| Generic BERT | 95.43% | 97.32% | 96.80% | 0.974 |
| Legal-BERT (our approach) | 96.82% | 98.93% | 98.25% | 0.984 |
| Rule-based | 98.76% | 68.42% | 77.32% | 0.715 |

## V. Conclusion and Future Work

This paper presented a comprehensive approach to definition extraction from the United States Code using transformer-based models and hierarchical document understanding. Our system demonstrates significant improvements over previous approaches by leveraging legal domain adaptation, document structure awareness, and multi-task learning for joint extraction of definitions, terms, and scope.

### A. Key Contributions

The key contributions of this work include:

1) A robust methodology for processing hierarchical legal documents that preserves structural relationships critical for definition understanding
2) A transformer-based approach using legal domain adaptation that significantly outperforms traditional classification methods
3) A multi-task learning framework that jointly optimizes definition detection, term extraction, and scope identification
4) Comprehensive evaluation across multiple titles of the U.S. Code, demonstrating generalizability across different legal domains

### B. Limitations

Despite the strong performance of our system, several limitations remain:

1) Implicit definitions without standard definitional keywords remain challenging to identify reliably
2) Complex scope references, particularly those that refer to non-adjacent sections, can be difficult to resolve accurately
3) The current approach does not fully capture the temporal aspects of definitions, such as amendments and effective dates
4) Cross-references between definitions create dependencies that are not fully modeled in the current system

5) The performance degradation on multi-paragraph definitions indicates ongoing challenges with long-range contextual understanding

### C. Future Work

This work opens up several directions for future research

1) **Tracking Definition Changes**: Developing methods to track changes in definitions across different versions of the U.S. Code, enabling analysis of legal concept evolution over time.
2) **Analyzing Definition Networks**: Expanding our approach to create networks that show how legal terms are connected, helping to understand the structure of legal language.
3) **Comparing Across Jurisdictions**: Extending the methodology to compare definitions across different jurisdictions (federal, state, international), identifying conflicts, similarities, and differences in legal terminology.
4) **Interactive Definition Exploration**: Developing user interfaces that allow legal professionals to explore definitional relationships through visualization and navigation tools, improving accessibility of complex legal concepts.

The methodology presented in this paper provides a foundation for these future directions while delivering immediate practical value for legal information systems. By making legal definitions more accessible and understandable, this work contributes to broader goals of improving legal transparency and access to justice through computational methods.